\documentclass[11pt,a4paper]{article}
\usepackage[hyperref]{emnlp2018}
\usepackage{times}
\usepackage{latexsym}

\usepackage{booktabs}
\usepackage{bm}
\usepackage{tikz}
\usepackage{pgfplots}
\usepackage{graphicx}
	\usepackage{siunitx}
\DeclareSIUnit\eur{\officialeuro}
\DeclareSIUnit\M{M}
\DeclareSIUnit\k{k}

\usepackage{url}
\usepackage{caption}
\usepackage{amsmath}
\usepackage{amsfonts}
\usepackage{amssymb}
\usepackage{xparse}
\NewDocumentCommand{\ceil}{s O{} m}{%
	\IfBooleanTF{#1} 
	{\left\lceil#3\right\rceil} 
	{#2\lceil#3#2\rceil} 
}
\newcommand{\norm}[1]{\left\lVert#1\right\rVert}

\pgfplotsset{
	ylabel right/.style={
		after end axis/.append code={
			\node [rotate=90, anchor=north] at (rel axis cs:1,0.5) {#1};
		}   
	}
}

\aclfinalcopy 
\def\aclpaperid{940} 

\title{Adaptive Document Retrieval for Deep Question Answering}

\author{Bernhard Kratzwald {\normalfont  and} Stefan Feuerriegel \\
  Chair of Management Information Systems \\
  ETH Zurich \\
  Zurich, Switzerland \\
  {\tt \{bkratzwald, sfeuerriegel\}@ethz.ch} \\
}

\begin{document}
\maketitle
\begin{abstract}
State-of-the-art systems in deep question answering proceed as follows: (1)~an initial document retrieval selects relevant documents, which (2)~are then processed by a neural network in order to extract the final answer. Yet the exact interplay between both components is poorly understood, especially concerning the number of candidate documents that should be retrieved. We show that choosing a static number of documents -- as used in prior research -- suffers from a noise-information trade-off and yields suboptimal results. As a remedy, we propose an adaptive document retrieval model. This learns the optimal candidate number for document retrieval, conditional on the size of the corpus and the query. We report extensive experimental results showing that our adaptive approach outperforms state-of-the-art methods on multiple benchmark datasets, as well as in the context of corpora with variable sizes.
\end{abstract}

\section{Introduction}


Question-answering~(QA) systems proceed by following a two-staged process \cite[][]{Belkin.1993}: in a first step, a module for document retrieval selects $n$ potentially relevant documents from a given corpus. Subsequently, a machine comprehension module extracts the final answer from the previously-selected documents. The latter step often involves hand-written rules or machine learning classifiers \cite[c.\,f.][]{Shen.2006,Kaisser.2004}, and recently also deep neural networks \cite[e.\,g.][]{Chen.2017,Wang.2018}


The number of candidate documents $n$ affects the interplay between both document retrieval and machine comprehension component. A larger $n$ improves the recall of document retrieval and thus the chance of including the relevant information. However, this also increases the noise and might adversely reduce the accuracy of answer extraction. It was recently shown that a top-$1$ system can potentially outperform a system selecting more than one document \cite{Kratzwald.2018}. This finding suggests that a static choice of $n$ can result a suboptimal performance.


\textbf{Contributions.} This work analyzes the interplay between document retrieval and machine comprehension inside neural QA systems. We first reason numerically why a fixed choice of $n$ in document retrieval can negatively affect the performance of question answering. We thus propose a novel machine learning model that adaptively selects the optimal $n_i$ for each document retrieval. The resulting system outperforms state-of-the-art neural question answering on multiple benchmark datasets. Notably, the overall size of the corpus affects the optimal $n$ considerably and, as a result, our system evinces as especially superior over a fixed $n$ in settings where the corpus size is unknown or grows dynamically. 

\section{Related Work}
\begin{figure*}[t!]
	\footnotesize
	\makebox[\textwidth]{
		\begin{tabular}{ccc}
			(a)~Exact matches with correct answer & (b)~Recall at top-$n$ & (c)~Avg. number of relevant documents \\
			\includegraphics[width=.25\textwidth]{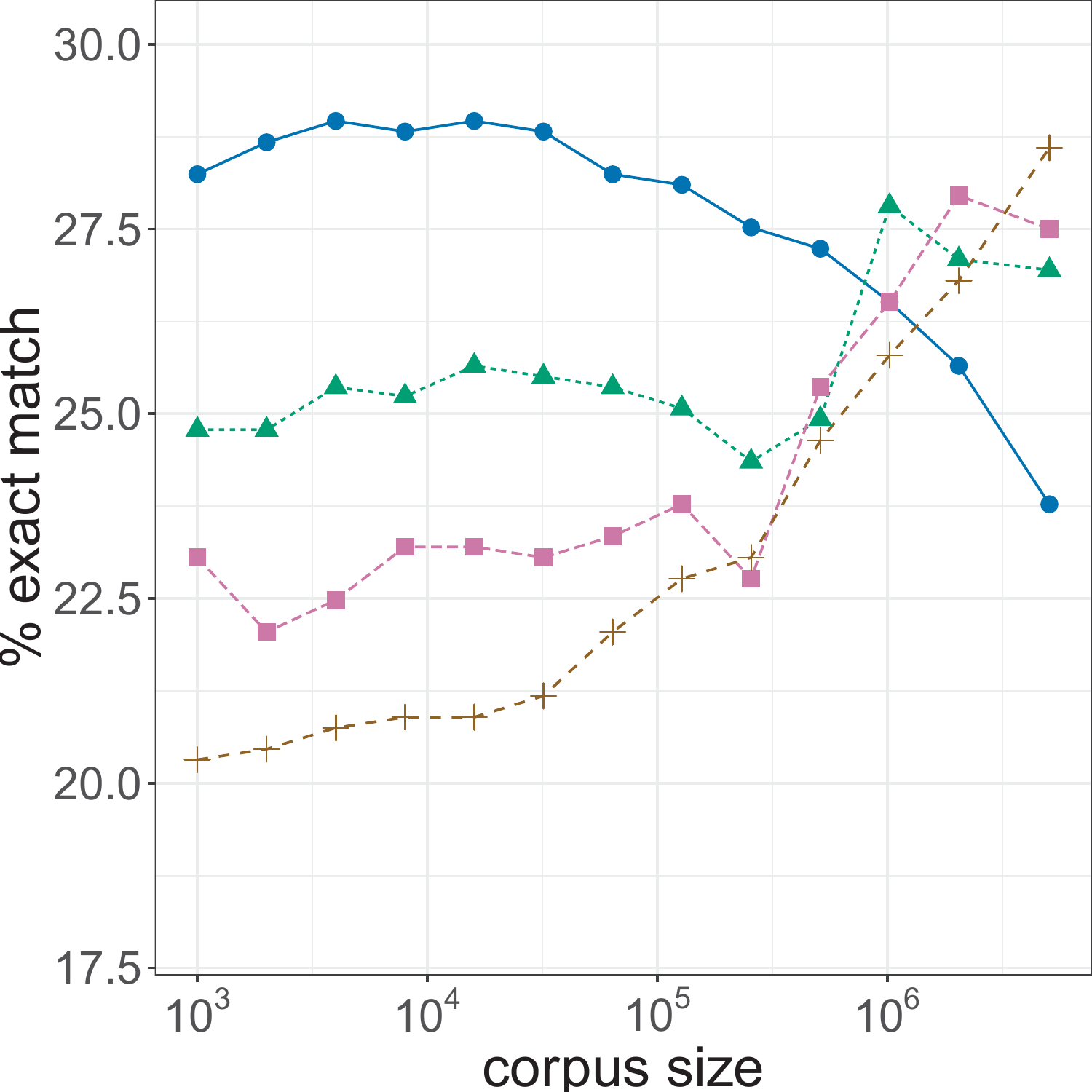} & \includegraphics[width=.25\textwidth]{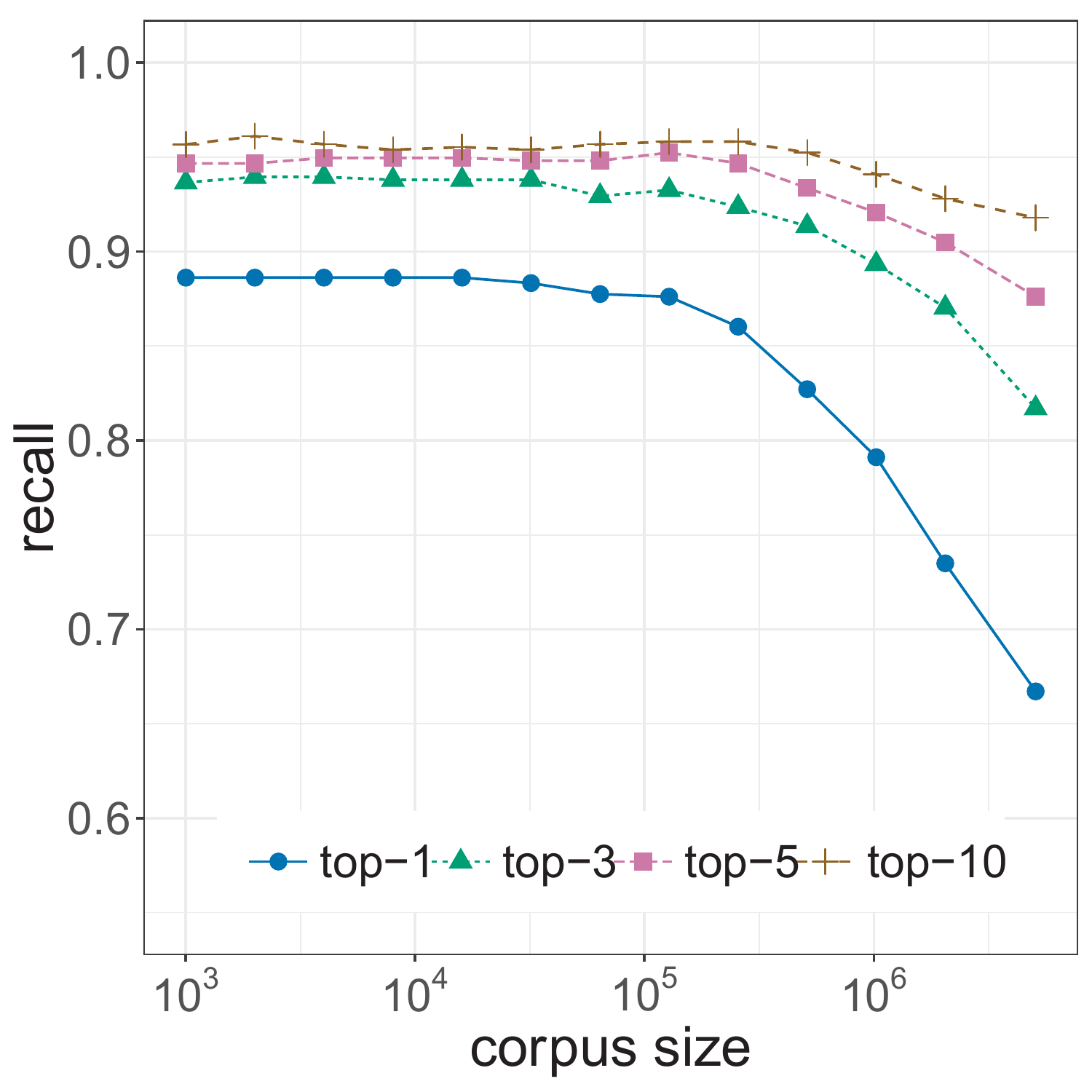} & 
			\includegraphics[width=.25\textwidth]{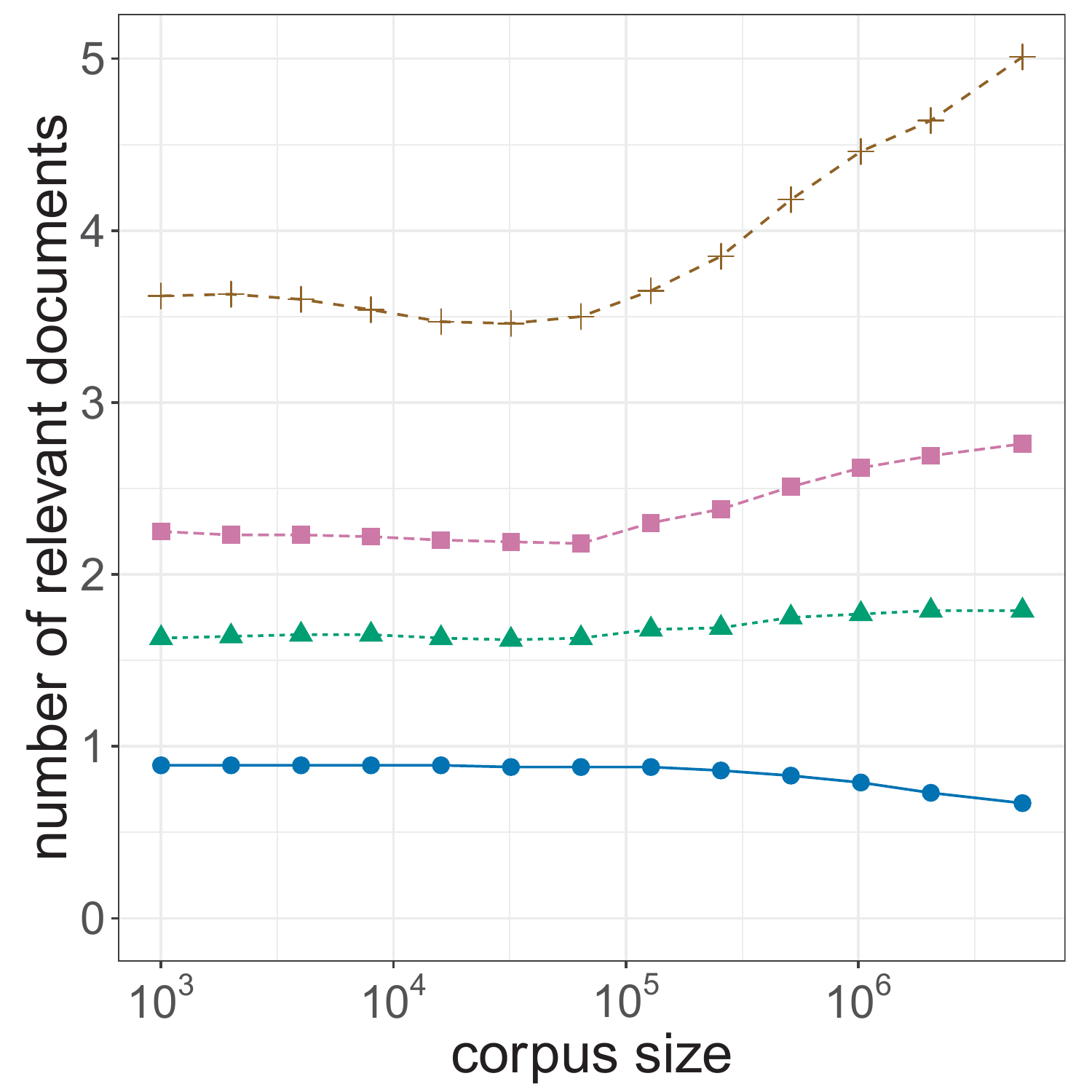}\\
		\end{tabular}
	}
	\caption{
		Comparison of how top-$n$ document retrieval affects deep QA. Plot~(a) shows the percentage of exact matches with the correct answering, thereby measuring the end-to-end performance of the complete system. Plot~(b) gives the recall at top-$n$, i.\,e. the fraction of samples where \emph{at least once} the correct answer is returned. Plot~(c) depicts the \emph{average number} of documents that contain the ground-truth answer. As a result, the recall lowers with increasing corpus size, yet this not necessarily compromises a top-$n$ system, as it often contains the correct answer more than once.}
	\label{fig:varyingcorpus}
\end{figure*}


\textbf{Taxonomy of QA systems.} Question answering systems are frequently categorized into two main paradigms. On the one hand, knowledge-based systems draw upon manual rules, ontologies and large-scale knowledge graphs in order to deduce answers~\cite[e.\,g.][]{Berant.2013, Lopez.2007, Unger.2012}. On the other hand, QA system incorporate a document retrieval module which selects candidate documents based on a chosen similarity metric, while a subsequent module then processes these in order to extract the answer~\cite[e.\,g.][]{Cao.2011,Harabagiu.2000}.


\textbf{Deep QA.} Recently, \citet{Chen.2017} developed a state-of-the-art deep QA system, where the answer is extracted from the top $n = 5$ documents. This choice stems from computing the dot product between documents and a query vector; with tf-idf weighting of hashed bi-gram counts. \citet{Wang.2018} extended this approach by implementing a neural re-ranking of the candidate document, yet keeping the fixed number of $n$ selected documents unchanged. In particular, the interplay between both modules for document retrieval and machine comprehension has not yet been studied. This especially pertains to the number of candidate documents, $n$, that should be selected during document retrieval.  


\textbf{Component interactions.} Extensive research has analyzed the interplay of both document retrieval and machine comprehension in the context of knowledge-based systems \cite[c.\,f.][]{Moldovan.2003} and even retrieval-based systems with machine learning \cite[c.\,f.][]{Brill.2002}. However, these findings do not translate to machine comprehension with deep learning. Deep neural networks consist of a complex attention mechanism for selecting the context-specific answer \cite{Hermann.2015} that has not been available to traditional machine learning and, moreover, deep learning is highly sensitive to settings involving multiple input paragraphs, often struggling with selecting the correct answer~\cite{Clark.2017}.

\section{Noise-Information Trade-Off in Document Retrieval}
\label{sec:sensitivity}


In the following, we provide empirical evidence why a one-fits-all $n$ can be suboptimal. For this purpose, we run a series of experiments in order to obtain a better understanding of the interplay between document retrieval and machine comprehension modules. That is, we specifically compare the recall of document retrieval to the end-to-end performance of the complete QA system; see Fig.~\ref{fig:varyingcorpus}. Our experiments study the sensitivity along two dimensions: on the one hand, we change the number of top-$n$ documents that are returned during document retrieval and, on the other hand, we vary the corpus size.


Our experiments utilize the TREC QA dataset as a well-established benchmark for open-domain question answering. It contains $694$ question-answer pairs that are answered with the help of Wikipedia. We vary the corpus between a small case (where each question-answer pair contains only one Wikipedia article with the correct answer plus \SI{50}{\percent} articles as noise) and the complete Wikipedia dump containing more than five million documents. Our experiments further draw upon the DrQA system \cite{Chen.2017} for question answering that currently stands as a baseline in deep question answering. We further modified it to return different numbers of candidate documents. 


Fig.~\ref{fig:varyingcorpus}~(a) shows the end-to-end performance across different top-$n$ document retrievals as measured by the exact matches with ground truth. For a small corpus, we clearly register a superior performance for the top-1 system. However, we observe a different pattern with increasing corpus size. Fig.~\ref{fig:varyingcorpus}~(b) and (c) shed light into the underlying reason by reporting how frequently the correct answer is returned and, as the correct answer might appear multiple times, how often it is included in the top-$n$. Evidently, the recall in~(b) drops quickly for a top-1 system when augmenting the corpus. Yet it remains fairly stable for a top-$n$ system, due to the fact that it is sufficient to have the correct answer in any of the $n$ documents. According to~(c), the correct answer is often more than once returned by a top-$n$ system, increasing the chance of answer extraction. 

\begin{figure}[t]
	\footnotesize
	\makebox[.5\textwidth]{
		\begin{tabular}{cc}
			(a)~Recall&(b)~Avg number of \\&relevant documents\\
			\includegraphics[width=.20\textwidth]{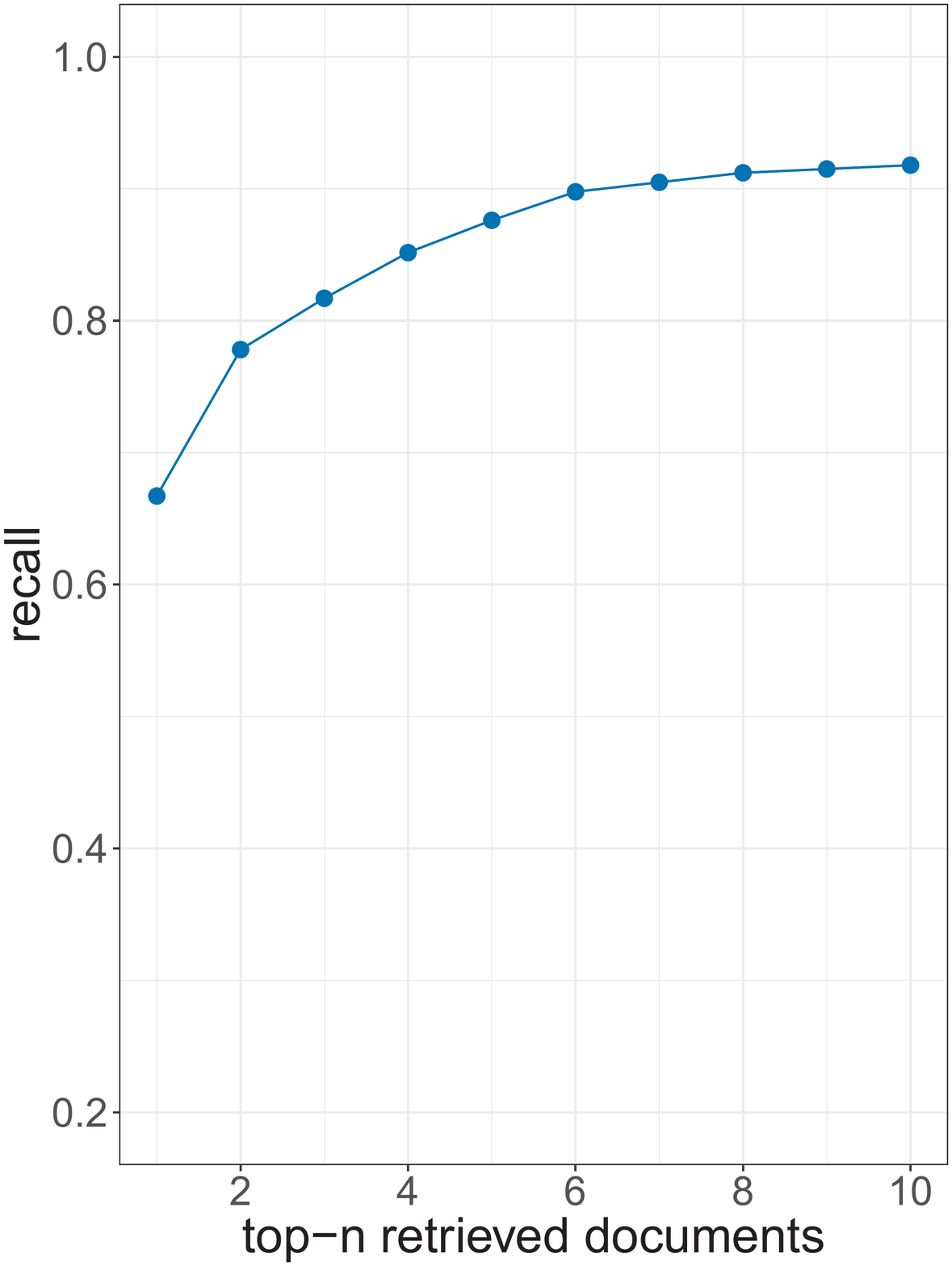} & \includegraphics[width=.20\textwidth]{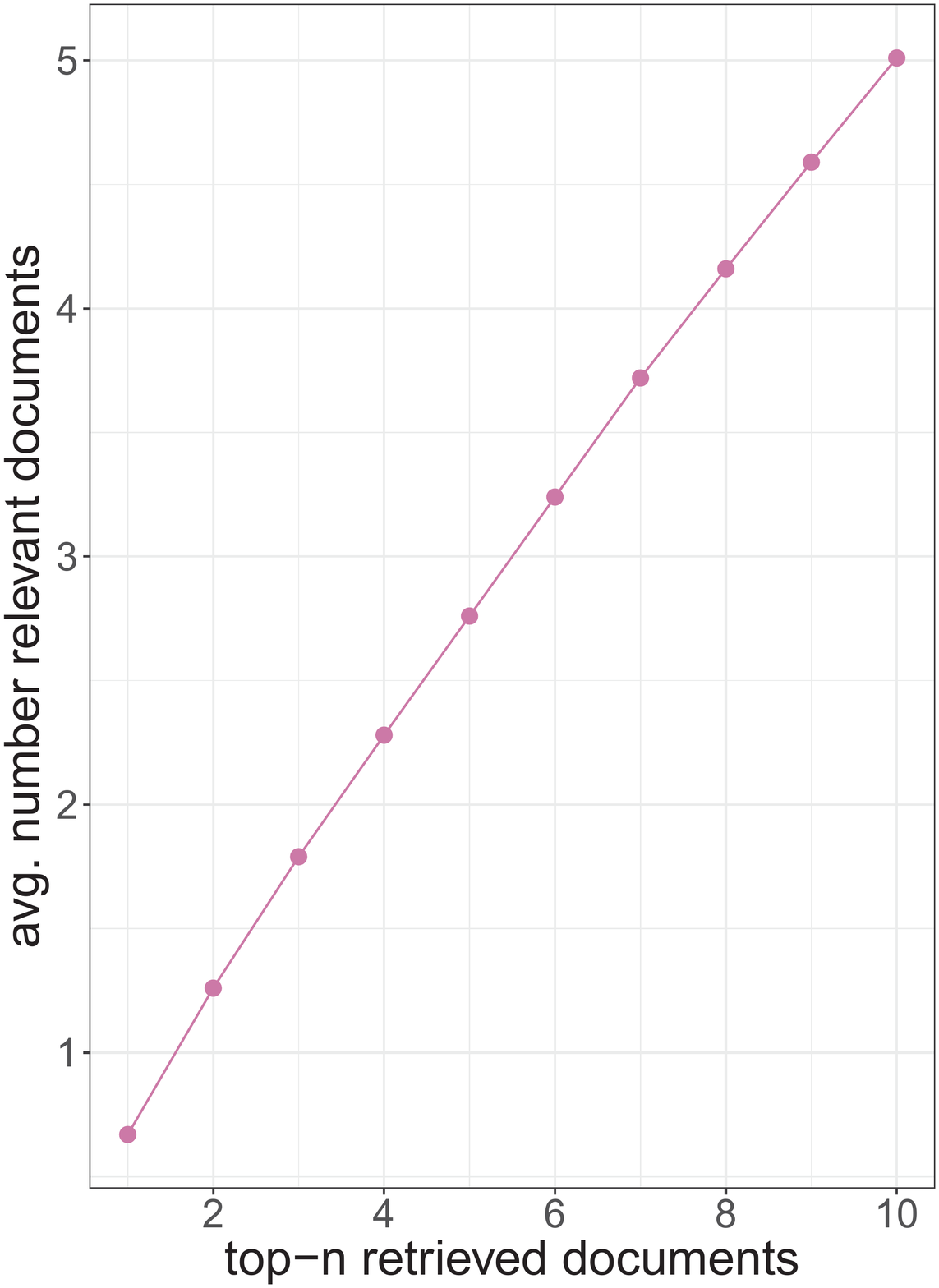} \\
		\end{tabular}
	}
	\caption{Recall (a) and average number of relevant documents (b) for growing top-$n$ configurations and a static corpus size (full Wikipedia dump). While the recall is converging the number of relevant documents keeps growing resulting in a higher density of relevant information.}
	\label{fig:density}
\end{figure}

The above findings result in a noise-information trade-off. A top-$1$ system often identifies the correct answer for a small corpus, whereas a larger corpus introduces additional noise and thus impedes the overall performance. Conversely, a \mbox{top-$n$} system accomplishes a higher density of relevant information for a large corpus as the answer is often contained multiple times. This effect is visualized in an additional experiment shown in Fig.~\ref{fig:density}. We keep the corpus size fixed and vary only $n$, i.e. the number of retrieved documents. We see the recall converging fast, while the average number of relevant documents keeps growing, leading to a higher density of relevant information. As a result, a top-$n$ system might not be compromised by a declining recall, since it contains the correct answer over-proportionally often. This logic motivates us in the following to introduce an adaptive $n_i$ that optimizes the number of documents retrievals in a top-$n$ system independently for every query $q_i$.

\section{Adaptive Document Retrieval}

This section advances deep question answering by developing adaptive methods for document retrieval. Our methods differ from conventional document retrieval in which the number of returned documents is set to a fixed $n$. Conversely, we actively optimize the choice of $n_i$ for each document retrieval $i$. Formally, we select $n_i$ between $1$ and a maximum $\tau$ (e.\,g. $\tau=20$), given documents $[d_i^{(1)}, \ldots, d_i^{(\tau)}]$. These entail further scores denoting the relevance, i.\,e. $s_i = [s_i^{(1)}, \ldots, s_i^{(\tau)}]^T$ with normalization s.\,t. $\sum_j{s_i^{(j)}}=1$. The scoring function is treated as a black-box and thus can be based on simple tf-idf similarity but also complex probabilistic models.

\subsection{Threshold-Based Retrieval}

As a na{\"i}ve baseline, we propose a simple threshold-based heuristic. That is, $n_i$ is determined such that the cumulative confidence score reaches a fixed threshold $\theta \in \left(0,1\right]$. Formally, the number $n_i$ of retrieved documents is given by 
\begin{equation}
n_i = \max_{k}{\sum_{j=1}^k{s_i^{(j)}} < \theta}.
\end{equation}
In other words, the heuristic fills up documents until surpassing a certain confidence threshold. For instance, if the document retrieval is certain that the correct answer must be located within a specific document, it automatically selects fewer documents. 
	
\subsection{Ordinal Regression}
We further implement a trainable classifier in the form of an ordinal ridge regression which is tailored to ranking tasks. We further expect the cumulative confidence likely to be linear. The classifier then approximates $n_i$ with a prediction $y_i$ that denotes the position of the first relevant document containing the desired answer. As such, we learn a function 
\begin{equation}
y_i = f([s_i^{(1)}, \ldots, s_i^{(\tau)}]) = \ceil{s_i^T\beta},
\end{equation} 
where $\ceil{\ldots}$ denotes the ceiling function.

The ridge coefficients are learned through a custom loss function 
\begin{equation}
\mathcal{L} = \norm{\ceil{X\beta}-y}_1 + \lambda\norm{\beta}_2 ,
\end{equation} 
where $X$ is a matrix containing scores of our training samples. In contrast to the classical ridge regression, we introduce a ceiling function and replace the mean squared error by a mean absolute error in order to penalize the difference from the optimal rank. The predicted cut-off $\hat{n}_i$ for document retrieval is then computed for new observations ${s'}_i$ via 
$
	\hat{n}_i = \ceil{{s'}_i^T \hat{\beta}} + b
$. 
 The linear offset $b$ is added in order to ensures that $n_i \leq \hat{n}_i$ holds, i.\,e. reducing the risk that the first relevant document is not included.

We additionally experimented with non-linear predictors, including random forests and feed-forward neural networks; however; we found no significant improvement that justified the additional model complexity over the linear relationship. 

\begin{figure*}[h!]
	\footnotesize
	\begin{minipage}[T!]{0.28\textwidth}
		\includegraphics[width=1\textwidth]{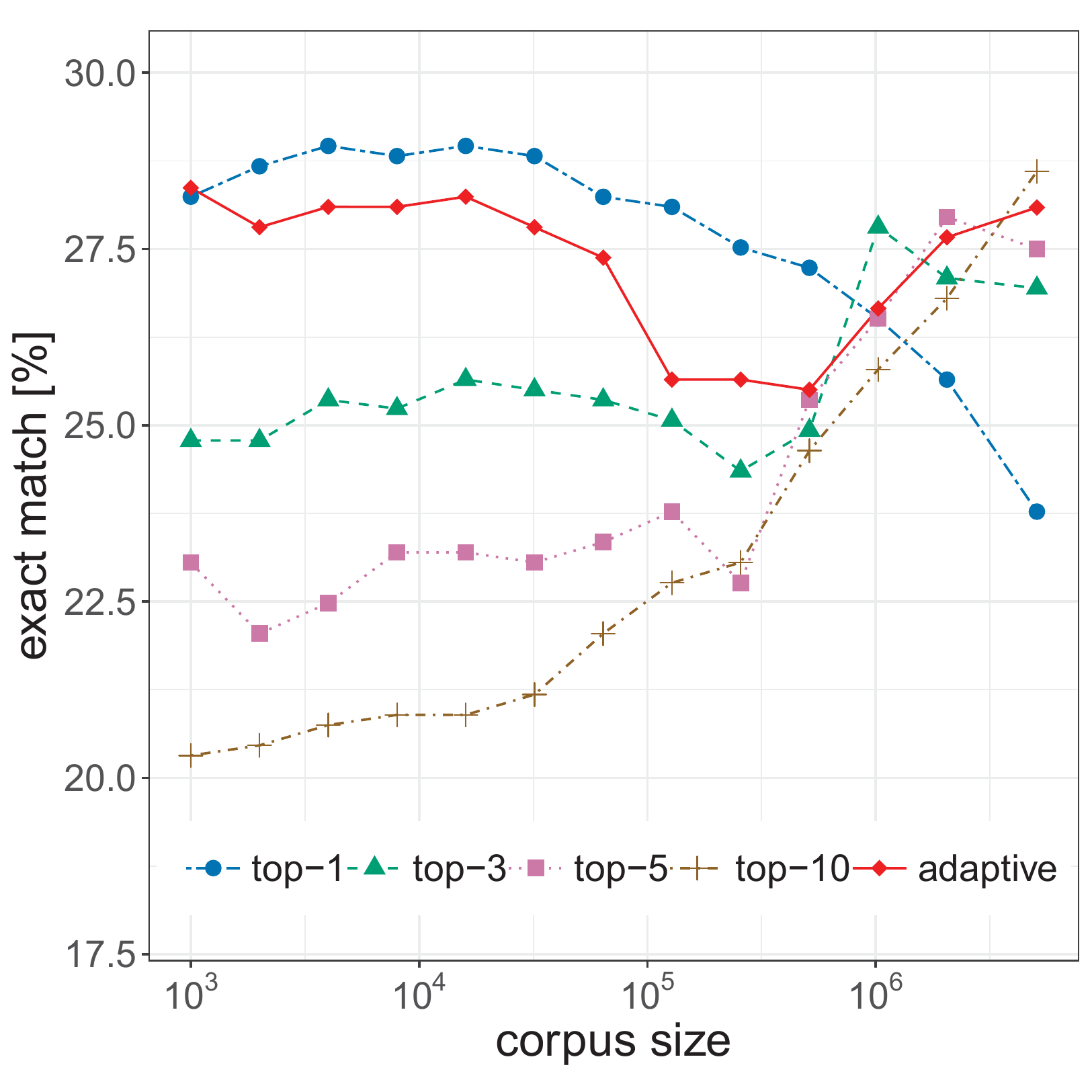}
		\captionof{figure}{End-to-end performance of adaptive information retrieval over static top-$n$ configurations and a growing corpus.}
		\label{fig:growingreg}
	\end{minipage}
	\hfill
	\begin{minipage}[T!]{0.70\textwidth}
		\begin{tabular}{l cccc}
			\toprule
			& {\bf SQuAD} & {\bf TREC} & {\bf WebQuestions} & {\bf WikiMovies}\\
			\midrule
			DrQA~\cite{Chen.2017}$^\dagger$ & 29.3		   & 27.5 & 18.5 & 36.6 \\
			\midrule
			Threshold-based  ($\theta = 0.75$)  & \bfseries 29.8 &  28.7 & 19.2 			 &  38.6 \\
			\midrule
			Ordinal regression ($b = 1$) & 29.7 & 28.1 & 19.4 & 38.0 \\
			Ordinal regression ($b = 3$) & 29.6 &\bfseries 29.3 & \bfseries 19.6 & 38.4 \\
			\midrule
			\midrule
			$R^3$~\cite{Wang.2018} & 29.1 & 28.4 & 17.1 & \bfseries38.8 \\
			\bottomrule
			\multicolumn{5}{l}{\tiny $^\dagger$:~Numbers vary slightly from those reported in the original paper, as the public repository was optimized for runtime performance.}
		\end{tabular} 
		\captionof{table}{End-to-end performance of the plain DrQA system measured in exact matches. Performance of two threshold based and two regression based adaptive retreival improvements as well as other state-of-the art systems. Experiments are based on the full Wikipedia dump containing more than 5 million documents.} 
		\label{tab:sensitivity}
	\end{minipage}	
\end{figure*}

\section{Experiments}

 
We first compare our QA system with adaptive document retrieval against benchmarks from the literature. Second, we specifically study the sensitivity of our adaptive approach to variations in the corpus size.  All our experiments draw upon the DrQA implementation \cite{Chen.2017}, a state-of-the-art system for question answering in which we replaced the default module for document retrieval with our adaptive scheme (but leaving all remaining components unchanged, specifically without altering the document scoring or answer extraction).

For the threshold-based model, we set $\tau=15$ and the confidence threshold to $\theta = 0.75$. For the ordinal regression approach, we choose $\tau=20$ and use the original SQuAD train-dev split from the full corpus also as the basis for training across all experiments. 

\subsection{Overall Performance}


In a first series of experiments, we refer to an extensive set of prevalent benchmarks for evaluating QA systems, namely, SQuAD~\cite{Rajpurkar.2016}, Curated TREC~\cite{Baudis.2015}, WikiMovies~\cite{Miller.2016} and WebQuestions~\cite{Berant.2013} in order to validate the robustness of our findings. Based on these, we then evaluate our adaptive QA systems against the na{\"i}ve DrQA system in order to evaluate the relative performance. We included the deep QA system $R^3$ as an additional, top-scoring benchmark from recent literature \cite{Wang.2018} for better comparability.


Tbl.~\ref{tab:sensitivity} reports the ratio of exact matches for the different QA systems. The results demonstrate the effectiveness of our adaptive scheme: it yields the best-performing system for three out of four datasets. On top of that, it outperforms the na{\"i}ve DrQA system consistently across all datasets.

\begin{table*}[h]
\footnotesize
\centering
	\begin{tabular}{l cccc}
		\toprule
		& {\bf SQuAD} & {\bf TREC} & {\bf WebQuestions} & {\bf WikiMovies}\\
		\midrule
		Top-$50$ System & 27.0	   & 23.5 & 15.1 & 24.4 \\
		Top-$80$ System & 27.2	   & 25.9 & 14.9 & 26.0 \\
		\midrule
		Threshold-based ($\theta = 0.75, \tau=100$) & 27.2 &\bfseries 27.1 & 15.4&26.3 \\
		Ordinal regression ($b=3, \tau=250$)  &\bfseries  27.3 & \bfseries 27.1 & \bfseries16.7 &\bfseries 26.5 \\
		\bottomrule
	\end{tabular} 

\caption{End-to-end performance measured in percentages of exact matching answers of a second QA system that operates on paragraph-level information retrieval. We compare two configurations of the system using the top-$50$ and top-$80$ ranked paragraphs to extract the answer against our threshold-based approach and regression approach that selects the cutoff within the first $250$ paragraphs.}
\label{tab:second-qa}
\end{table*}

\subsection{Sensitivity: Adaptive QA to Corpus Size} 


We earlier observed that the corpus size affects the best choice of $n$ and we thus study the sensitivity with regard to the size. For this purpose, we repeat the experiments from Section~\ref{sec:sensitivity} in order to evaluate the performance gain from our adaptive scheme. More precisely, we compare the ordinal regression ($b = 1$) against document retrieval with a fixed document count $n$.  


Fig.~\ref{fig:growingreg} shows the end-to-end performance, confirming the overall superiority of our adaptive document retrieval. For instance, the top-1 system reaches a slightly higher rate of exact matches for small corpus sizes, but is ranked last when considering the complete corpus. The high performance of the top-$1$ system partially originates from the design of the experiment itself, where we initially added \emph{one} correct document per question, which is easy to dissect by adding little additional noise. On the other hand, the top-10 system accomplishes the best performance on the complete corpus, whereas it fails to obtain an acceptable performance for smaller corpus sizes.

To quantify our observations, we use a notation of regret. Formally, let $\mu_{nm}$ denote the performance of the top-$n$ system on a corpus of size~$m$. Then the regret of choosing system $n$ at evaluation point~$m$ is  the difference between the best performing system $\mu^*_m$ and the chosen system \mbox{$r_{nm} =  \mu^*_m - \mu_{nm}$}. The total regret of system $n$ is computed by averaging the regret over all observations of system~$n$, weighted with the span in-between observations in order to account for the logarithmic intervals. The best top-$n$ system yields a regret of $0.83$ and $1.12$ respectively, whereas our adaptive control improves it down to $0.70$.

\subsection{Robustness Check}
Experiments so far have been conducted on the DrQA system. To show the robustness of our approach, we repeat all experiments on a different QA system. Different from DrQA, this system operates on paragraph-level information retrieval and uses cosine similarity to score tf-idf-weighted bag-of-word (unigram) vectors. The reader is a modified version of the DrQA document reader with an additional bi-directional attention layer~\cite{Seo.2016}. We are testing two different configurations\footnote{Best configurations out of $\{30,40,50,60,70,80,$ $90,\text{\,and\,}100\}$ on  SQuAD train split.} of this system: one that selects the top-$50$ paragraphs and one that selects the top-$80$ paragraphs against our approach as shown in Tab.~\ref{tab:second-qa}. We see that, owed to the paragraph-level information retrieval, the number of top-$n$ passages gains even more importance. Both variations of the system outperform a system without adaptive retrieval, which confirms our findings.

\section{Conclusion}

Our contribution is three-fold. First, we establish that deep question answering is subject to a noise-information trade-off. As a consequence, the number of selected documents in deep QA should not be treated as fixed, rather it must be carefully tailored to the QA task. Second, we propose adaptive schemes that determine the optimal document count. This can considerably bolster the performance of deep QA systems across multiple benchmarks. Third, we further demonstrate how crucial an adaptive document retrieval is in the context of different corpus sizes. Here our adaptive strategy presents a flexible strategy that can successfully adapt to it and, compared to a fixed document count, accomplishes the best performance in terms of regret. 

\section*{Reproducibility}
Code to integrate adaptive document retrieval in custom QA system and future research is freely available at \url{https://github.com/bernhard2202/adaptive-ir-for-qa}

\section*{Acknowledgments}
We thank the anonymous reviewers for their helpful comments. We gratefully acknowledge the support of NVIDIA Corporation with the donation of the Titan Xp GPU used for this research. Cloud computing resources were provided by a Microsoft Azure for Research award.

\bibliography{refs}
\bibliographystyle{acl_natbib_nourl}

\end{document}